\documentclass[conference]{IEEEtran}
\IEEEoverridecommandlockouts

\usepackage{cite}
\usepackage{amsmath,amssymb,amsfonts}
\usepackage{algorithmic}
\usepackage{graphicx}
\usepackage{textcomp}
\usepackage{xcolor}
\def\BibTeX{{\rm B\kern-.05em{\sc i\kern-.025em b}\kern-.08em
    T\kern-.1667em\lower.7ex\hbox{E}\kern-.125emX}}

\usepackage{url}
\usepackage{array}
\usepackage{caption}
\usepackage{subcaption}
\usepackage{algorithm}
\usepackage{algorithmic}
\usepackage{cleveref}
\usepackage{tikz}
\usepackage{booktabs}
\usetikzlibrary{shapes.geometric, arrows}
\usepackage{pgfplots}

\begin{document}

\title{Causal Time Series Modeling of Supraglacial Lake Evolution in Greenland under Distribution Shift\\


}

\author{
    \IEEEauthorblockN{
        Emam Hossain\IEEEauthorrefmark{1},
        Muhammad Hasan Ferdous\IEEEauthorrefmark{1},
        Devon Dunmire\IEEEauthorrefmark{2}, 
        Aneesh Subramanian\IEEEauthorrefmark{3},
        Md Osman Gani\IEEEauthorrefmark{1}
    }
    \IEEEauthorblockA{
        \IEEEauthorrefmark{1}Department of Information Systems, University of Maryland Baltimore County, USA \\
        \IEEEauthorrefmark{2}Department of Earth and Environmental Sciences, KU Leuven, Belgium\\
        \IEEEauthorrefmark{3}Department of Atmospheric and Oceanic Sciences, University of Colorado Boulder, USA \\
        Email: \textit{\{emamh1, h.ferdous, mogani\}@umbc.edu, devon.dunmire@kuleuven.be, aneeshcs@colorado.edu}
    }
}

\maketitle

\begin{abstract}
Causal modeling offers a principled foundation for uncovering stable, invariant relationships in time-series data, thereby improving robustness and generalization under distribution shifts. Yet its potential is underutilized in spatiotemporal Earth observation, where models often depend on purely correlational features that fail to transfer across heterogeneous domains. We propose RIC-TSC, a regionally-informed causal time-series classification framework that embeds lag-aware causal discovery directly into sequence modeling, enabling both predictive accuracy and scientific interpretability. Using multi-modal satellite and reanalysis data—including Sentinel-1 microwave backscatter, Sentinel-2 and Landsat-8 optical reflectance, and CARRA meteorological variables—we leverage Joint PCMCI+ (J-PCMCI+) to identify region-specific and invariant predictors of supraglacial lake evolution in Greenland. Causal graphs are estimated globally and per basin, with validated predictors and their time lags supplied to lightweight classifiers. On a balanced benchmark of 1000 manually labeled lakes from two contrasting melt seasons (2018–2019), causal models achieve up to 12.59\% higher accuracy than correlation-based baselines under out-of-distribution evaluation. These results show that causal discovery is not only a means of feature selection but also a pathway to generalizable and mechanistically grounded models of dynamic Earth surface processes.
\end{abstract}


\section{Introduction}

The Greenland Ice Sheet (GrIS) has experienced accelerating mass loss over recent decades, now contributing approximately one-quarter of observed global sea-level rise~\cite{shepherd2020mass}. A major driver of this mass loss is surface meltwater runoff, much of which is regulated by the seasonal formation and evolution of supraglacial lakes (SGLs)~\cite{van2016recent}. These lakes form when surface meltwater accumulates in topographic depressions during the summer, and their eventual fate—drainage, burial, or refreezing—can significantly influence downstream meltwater routing, firn structure, and basal sliding~\cite{das2008fracture}. The precise pathway a lake follows has implications for local ice dynamics and broader climate feedbacks.

\begin{figure}[!h]
    \centering
    \includegraphics[width=0.42\textwidth]{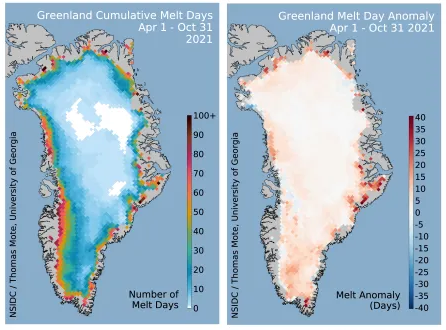}
    \caption{Cumulative melt days (left) and melt day anomaly relative to the 1981--2010 average (right) across the Greenland Ice Sheet for April--October 2021. (Source: NSIDC/Thomas Mote, University of Georgia)}
    \label{fig:grismelt}
\end{figure}

However, modeling SGL evolution remains a scientific challenge due to the spatially heterogeneous and temporally variable nature of melt conditions. Figure~\ref{fig:grismelt} illustrates both the cumulative extent of melting in 2021 and its anomaly relative to the 1981--2010 average, highlighting how melt intensity and spatial distribution have increased and shifted in recent decades. These patterns vary not only from year to year but also across Greenland’s six drainage basins—each with distinct climatological regimes. Consequently, lakes exposed to similar large-scale weather events can behave very differently depending on their regional context. Drainage events may trigger sudden basal lubrication and ice acceleration~\cite{zwally2002surface}, while burial and refreezing reshape firn structure and precondition future hydrological responses~\cite{macferrin2019rapid}.

Traditional machine learning approaches for SGL classification rely on handcrafted or correlated features derived from satellite time series~\cite{hossain2024time}. While effective in local settings, these methods often fail to generalize across spatial domains or climate regimes due to their lack of physical grounding and causal interpretability. Deep sequence models like LSTM-FCN~\cite{karim2018lstmfcn} can capture temporal dynamics but require extensive labeled data and provide limited insight into the underlying mechanisms driving lake evolution.

In this work, we introduce \textbf{RIC-TSC}, a \textbf{R}egionally-\textbf{I}nformed \textbf{C}ausal \textbf{T}ime \textbf{S}eries \textbf{C}lassification framework for supraglacial lake classification that explicitly addresses generalization under spatial and climatic distribution shifts. Our approach goes beyond combining existing tools by tailoring deep learning methods to environmental data constraints: we integrate causal discovery with scalable sequence transforms to achieve both interpretability and robustness. Specifically, we leverage multi-modal satellite observations (Sentinel-1 SAR, Sentinel-2 and Landsat-8 optical) and climate reanalysis fields (CARRA-West) to construct per-lake daily time series for 1000 manually labeled lakes across six basins and two melt seasons (2018–2019). Using Joint PCMCI+ (J-PCMCI+)~\cite{runge2020discovering}, we estimate both global and region-specific causal graphs to identify temporally lagged drivers of lake evolution—variables that are not just correlated but statistically causal with respect to key lake dynamics such as backscatter anomaly (HV\textsubscript{anom}) and water percentage ($p_{\text{water}}$).

These lag-aware causal predictors are then provided as input to MiniROCKET~\cite{dempster2021minirocket}, a lightweight yet powerful time series transform that serves as a deep feature extractor, followed by RidgeClassifier for efficient sequence classification. This design balances speed, interpretability, and discriminative power while avoiding the overhead and opaqueness of more complex neural networks. Our contributions are as follows:
\begin{itemize}
    \item We present a novel causal time series classification framework called RIC-TSC for supraglacial lake evolution that integrates J-PCMCI+ causal discovery with deep sequence modeling via MiniROCKET.
    \item We construct global and basin-specific causal graphs from remote sensing and climate reanalysis data, identifying stable, interpretable predictors of lake outcomes.
    \item We classify lakes into four physically meaningful categories—\textit{refreeze}, \textit{buried}, \textit{rapid drainage}, and \textit{slow drainage}, using a pan-Greenland benchmark dataset spanning two contrasting melt seasons.
    \item We demonstrate that causal features substantially improve model generalization under both in-distribution and out-of-distribution settings, achieving up to +12.59\% accuracy gain in region-held-out evaluations.
\end{itemize}

By combining causal inference with deep learning-based sequence modeling, our framework advances the development of generalizable and interpretable tools for cryospheric science, offering new capabilities for understanding and forecasting supraglacial lake dynamics in a warming Greenland.

\section{Related Works}

\subsection{Causality in Machine Learning Models}

Generalization under distribution shift is a central challenge in machine learning, especially in domains with strong spatial or temporal variability. Causal inference addresses this by uncovering stable mechanisms that persist across environments~\cite{pearl2009causality}, unlike correlational models that often rely on spurious associations. Frameworks such as Invariant Causal Prediction (ICP)~\cite{peters2016causal} and Invariant Risk Minimization (IRM)~\cite{arjovsky2019invariant} identify feature subsets with predictive relationships invariant across environments, though they are less explored for multivariate time series.


Causal discovery in time series introduces additional challenges of autocorrelation, lags, and confounding. Granger causality~\cite{granger1969investigating} is widely used but assumes linearity and struggles under high dimensionality or hidden confounders. PCMCI~\cite{runge2019detecting} and PCMCI+~\cite{runge2020discovering} address these issues by combining constraint-based discovery with conditional independence testing to recover lagged causal graphs. J-PCMCI+ extends this by pooling across structured datasets and incorporating dummy variables (e.g., region or time), making it effective for environmental systems with spatial heterogeneity.

Beyond discovery, advances in causal representation learning~\cite{scholkopf2021towards} and meta-learning of structural causal models~\cite{bengio2020meta} show that embedding causal principles into deep learning pipelines improves interpretability and generalization. Yet most applications are limited to synthetic or controlled datasets, leaving open questions of scalability to complex, noisy, high-dimensional Earth observation data. In this work, we leverage J-PCMCI+ to identify lagged causal relationships in supraglacial lake time series and integrate them into sequence classification, moving beyond correlation-driven heuristics toward principled, interpretable modeling of cryospheric dynamics.

\subsection{Machine Learning for Earth Observation and Supraglacial Lakes}

Machine learning is increasingly central to Earth observation, supporting applications from land cover classification~\cite{zhu2017deep} to surface water mapping~\cite{yang2020combined} and glacier monitoring~\cite{kaab2005remote}. For supraglacial lakes (SGLs), remote sensing with optical and radar imagery has enabled large-scale monitoring of extent and dynamics, but many classification pipelines rely on static or per-image features with thresholding, clustering, or shallow supervised learning~\cite{moussavi2020antarctic}. These approaches often overlook temporal structure and struggle to generalize across regions or seasons.

To capture temporal dynamics, recent studies have explored time series classification of SGLs. Hossain et al.~\cite{hossain2024time} introduced a Reconstructed Phase Space (RPS) model with Gaussian Mixture Models (GMMs) to classify lakes into refreeze, drain, and buried categories using Sentinel-1 and Sentinel-2 data. While effective within individual regions, this approach remained correlational and required retraining due to limited transferability. Deep learning methods such as LSTM-FCN~\cite{karim2018lstmfcn} and temporal CNNs have also been applied, but these models are data-hungry, prone to overfitting, and difficult to interpret—issues that are especially limiting for SGL studies where labeled data are scarce and costly to obtain.

\begin{figure*}[!h]
    \centering
    \begin{subfigure}[t]{0.24\textwidth}
    \centering
    \includegraphics[width=\textwidth]{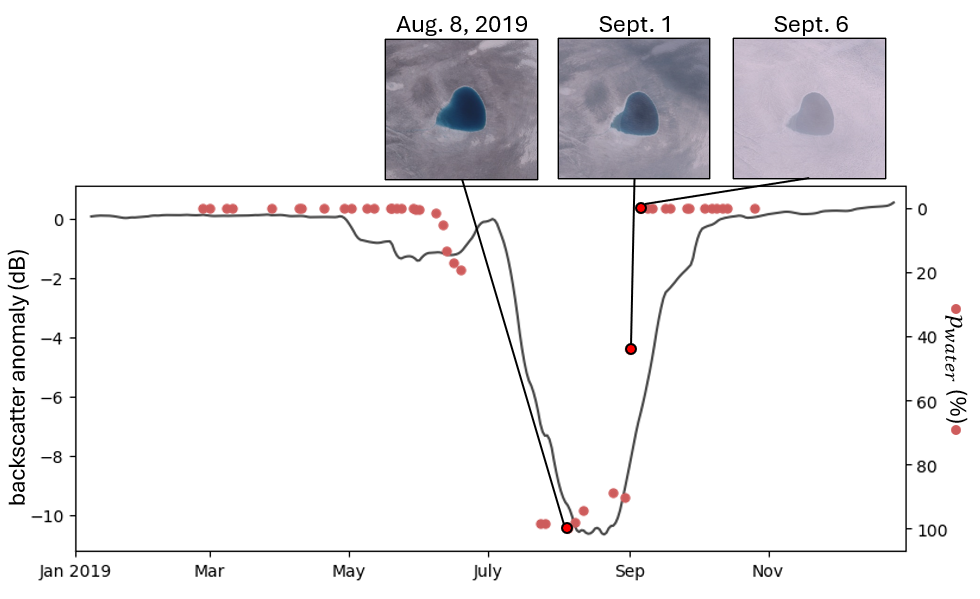}
    \caption{Refreeze}
    \label{fig:lake_types_refreeze}
    \end{subfigure}
    \hfill
    \begin{subfigure}[t]{0.24\textwidth}
    \centering
    \includegraphics[width=\textwidth]{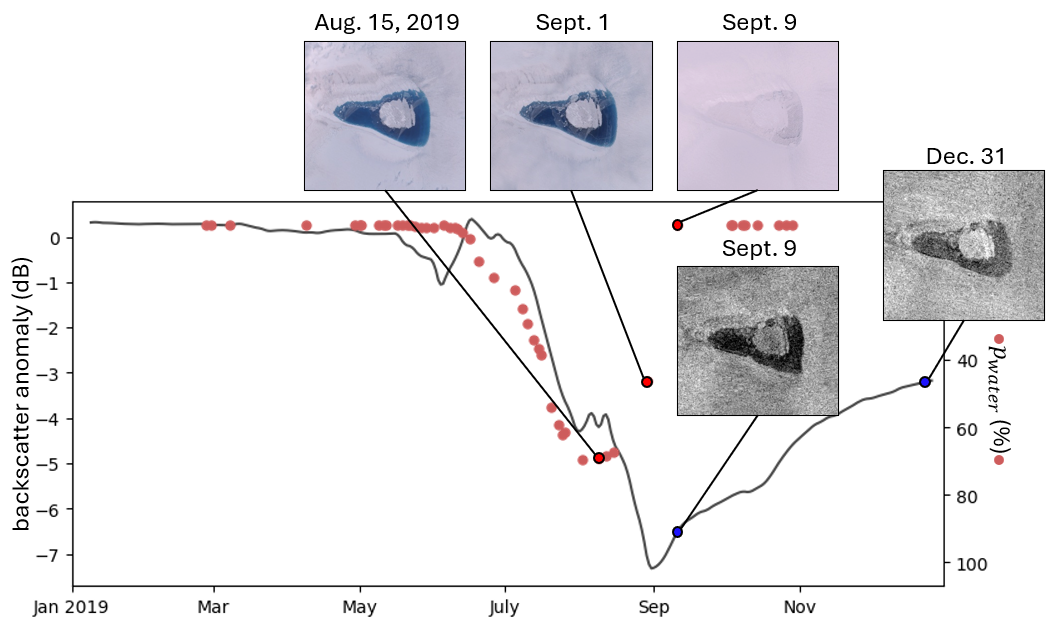}
    \caption{Buried}
    \label{fig:lake_types_buried}
    \end{subfigure}
    \hfill
    \begin{subfigure}[t]{0.24\textwidth}
    \centering
    \includegraphics[width=\textwidth]{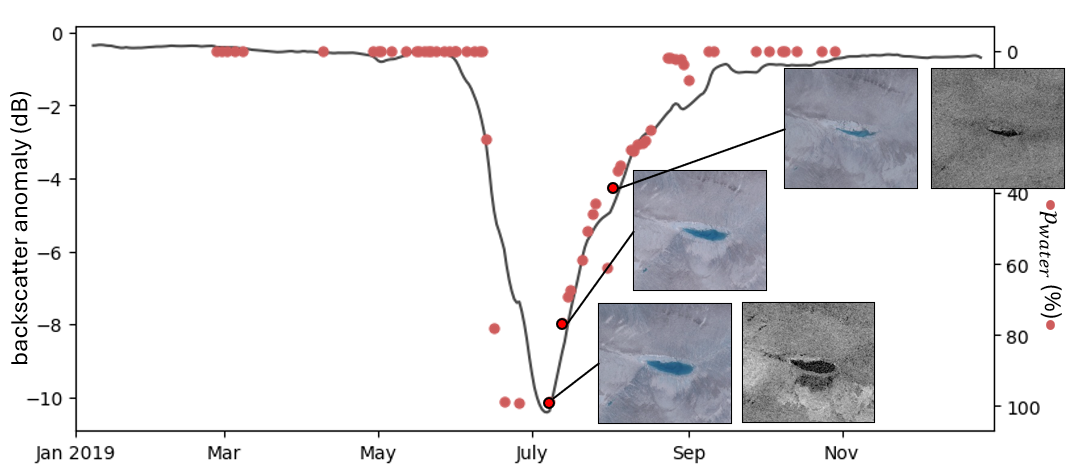}
    \caption{Slow drainage}
    \label{fig:lake_types_slowdrain}
    \end{subfigure}
    \hfill
    \begin{subfigure}[t]{0.24\textwidth}
    \centering
    \includegraphics[width=\textwidth]{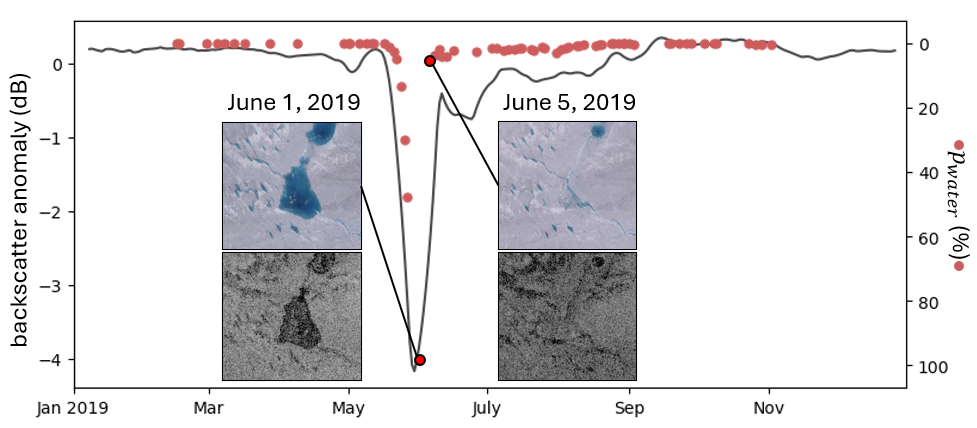}
    \caption{Rapid drainage}
    \label{fig:lake_types_rapiddrain}
    \end{subfigure}
    \caption{Representative time series and imagery for four lake evolution classes. Grey lines: backscatter anomaly ($HV_{\text{anom}}$). Red dots: optical water fraction ($p_{\text{water}}$). Insets: Sentinel-2 (color) and Sentinel-1 (grayscale) before/after the event. Modified and extended from \cite{hossain2024time}.}
    \label{fig:lake_types_all}
\end{figure*}

Few efforts have directly modeled the causal drivers of SGL evolution, even though ordered changes in temperature, solar zenith angle, and radar backscatter often indicate transitions such as drainage or burial. Incorporating such signals into a causal framework allows for more compact, interpretable, and generalizable models. Our work addresses this gap by combining causal discovery with efficient sequence classification: we use J-PCMCI+ to identify global and region-specific causal graphs and feed the resulting lag-aware predictors into a MiniROCKET + RidgeClassifier pipeline. This design improves generalization under distribution shifts while yielding insights into the dynamics governing lake evolution.

\section{Background}

\subsection{Supraglacial Lake Dynamics}

Supraglacial lakes (SGLs) form on the surface of the Greenland Ice Sheet during the summer melt season, driven by rising air temperatures, increased solar radiation, and surface runoff accumulation~\cite{sundal2009evolution}. These meltwater reservoirs play a central role in modulating the hydrological response of the ice sheet. Their evolution—whether through drainage, burial, or refreezing—can influence firn structure, basal lubrication, and surface meltwater routing, ultimately affecting regional ice dynamics and long-term mass balance~\cite{macferrin2019rapid}. Importantly, lake evolution is neither spatially uniform nor temporally consistent: even within the same basin, nearby lakes may follow divergent fates depending on topography, firn permeability, or climatic forcing. Traditional classifications group lakes into \textit{refreeze}, \textit{burial}, and \textit{drainage}, but recent studies reveal drainage to be a continuum, motivating a refined taxonomy. Following recent work on lake dynamics and time series classification~\cite{dunmire2021contrasting, hossain2024time}, we identify four distinct evolution pathways—\textbf{refreeze}, \textbf{buried}, \textbf{slow drainage}, and \textbf{rapid drainage}—each with different implications for meltwater retention, firn storage, and subglacial hydrology.

\begin{figure}[!h]
    \centering
    \includegraphics[width=0.40\textwidth]{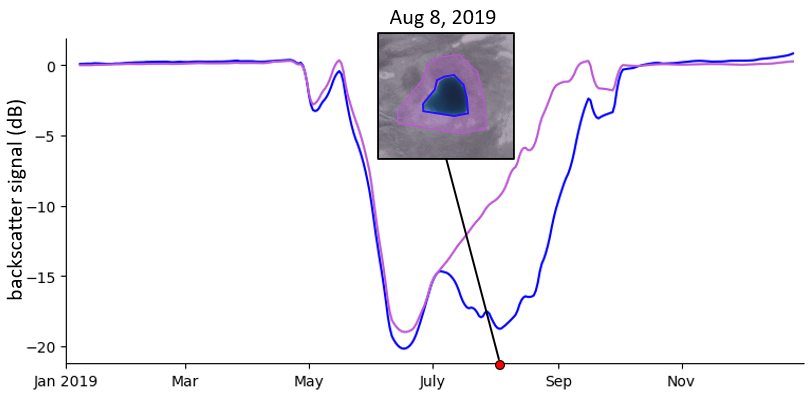}
    \caption{Comparison of radar backscatter signal from within a lake (blue) and its surroundings (purple). A large contrast indicates surface water; signal convergence suggests refreezing or burial. Modified and adapted from \cite{hossain2024time}.}
    \label{fig:hv_comparison}
\end{figure}

\paragraph{\textbf{Dual-Sensor Monitoring.}} We leverage Sentinel-1 C-band SAR and Sentinel-2 optical imagery to monitor lake evolution (Figure~\ref{fig:hv_comparison}). Radar backscatter is extracted from within the lake boundary ($HV_{\text{lake}}$) and a surrounding 750\,m buffer ($HV_{\text{out}}$). The anomaly is defined as:
\begin{equation}
HV_{\text{anom}} = HV_{\text{lake}} - HV_{\text{out}}
\label{eq:hv_anom}
\end{equation}
capturing the radar-dark signature of liquid water. In parallel, Sentinel-2 provides the water fraction:
\begin{equation}
p_{\text{water}} = \frac{N_{\text{water}}}{N_{\text{total}}} \times 100\%
\label{eq:p_water}
\end{equation}
where $N_{\text{water}}$ and $N_{\text{total}}$ are the number of water pixels and total pixels, respectively. Together, $HV_{\text{anom}}$ and $p_{\text{water}}$ offer complementary, temporally rich indicators of lake state.

\paragraph{\textbf{Lake Evolution Pathways.}} The four classes exhibit distinct dual-sensor signatures (Figure~\ref{fig:lake_types_all}). \textit{Refreeze lakes} (Figure~\ref{fig:lake_types_refreeze}) gradually solidify in place, with both $HV_{\text{anom}}$ and $p_{\text{water}}$ trending smoothly toward zero, indicating progressive freezing and long-term firn densification. \textit{Buried lakes} (Figure~\ref{fig:lake_types_buried}) appear optically dry as $p_{\text{water}} \to 0$, but persistently negative $HV_{\text{anom}}$ values reveal subsurface water insulated by snow or firn. \textit{Slow drainage lakes} (Figure~\ref{fig:lake_types_slowdrain}) lose water gradually over weeks, with both indicators decaying slowly as melt percolates into firn or shallow conduits, redistributing water internally. By contrast, \textit{Rapid drainage lakes} (Figure~\ref{fig:lake_types_rapiddrain}) collapse abruptly within 1–3 days through hydrofracture~\cite{das2008fracture}, observed as a sudden drop in $p_{\text{water}}$ and a sharp spike in $HV_{\text{anom}}$, directly routing meltwater to the ice-sheet base and driving transient flow acceleration. These empirically defined pathways ground our causal discovery and sequence modeling.

\subsection{Causal Discovery for Time Series}

\begin{figure}[!t]
    \centering
    \includegraphics[width=0.8\linewidth]{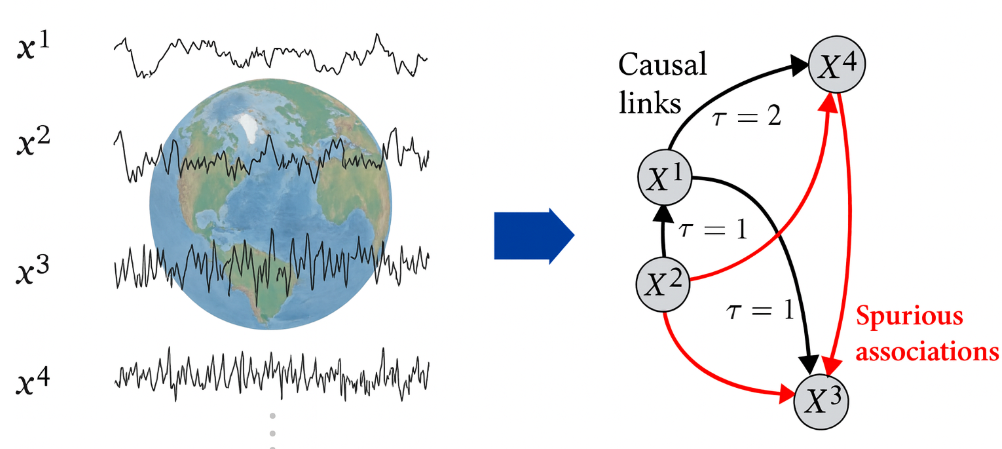}
    \caption{Causal discovery for multivariate time series. J-PCMCI+ recovers both lagged and contemporaneous causal links (black) while filtering spurious correlations (red), enabling interpretable modeling of SGL drivers across regions.}
    \label{fig:causal_discovery_process}
\end{figure}

Robust modeling of supraglacial lake dynamics requires identifying not only which variables matter but also the temporal lags at which they influence outcomes. Causal discovery methods aim to recover such directed, time-lagged dependencies (Figure~\ref{fig:causal_discovery_process}) that reflect underlying generative mechanisms rather than mere correlations~\cite{hasansurvey, ferdous2025timegraph, hossain2025learning, hossain2025correlation}. Classical Granger causality~\cite{granger1969investigating} formalized this in linear VAR models, but its assumptions of linearity, independence, and absence of hidden confounders limit applicability to noisy, nonlinear geophysical systems.

PCMCI~\cite{runge2019detecting} overcomes these issues using conditional independence tests, while PCMCI+~\cite{runge2020discovering} improves robustness by refining conditioning sets to detect both lagged and contemporaneous edges. J-PCMCI+ extends this to heterogeneous datasets by introducing structured dummy variables (e.g., basin, season, elevation), enabling pooled inference while still isolating stable causal mechanisms. 

In this study, we apply J-PCMCI+ to 365-day multivariate time series constructed from radar backscatter anomaly ($HV_{\text{anom}}$), optical water fraction (Sentinel-2, Landsat-8), solar zenith angle, and reanalysis drivers such as near-surface temperature (t2m), relative humidity (r2), surface pressure (sp), and sea surface temperature (sst). The resulting graphs identify both causal parents and their associated time lags—features that form the input space for downstream sequence classification. This ensures classification models rely on predictors with statistically causal, lag-aware influence on lake evolution.

\subsection{Sequence Modeling}

Once causal parents and their lags are identified, the task reduces to modeling temporal trajectories of lake states. Classical sequence models such as RNNs, GRUs, or LSTMs can capture nonlinear dependencies but are data-hungry, opaque, and prone to overfitting~\cite{islam2021foreign, dey2021comparative}—limitations in Earth observation where labeled datasets are small and heterogeneous. We instead adopt \textbf{MiniROCKET}~\cite{dempster2021minirocket}, a deterministic convolutional transform that generates discriminative features efficiently and is robust with limited data. Coupled with a \textbf{RidgeClassifier} for regularized linear decision-making, this provides a lightweight yet effective classification pipeline.

Crucially, inputs to MiniROCKET are not the full variable set but the causal parents with their discovered lags from J-PCMCI+. This hybrid design—causal discovery followed by efficient sequence modeling—enables parsimonious classification that leverages physically grounded, lag-aware predictors. As shown later, this improves generalization under both in-distribution and region-held-out settings, highlighting the stability of causal features for spatiotemporal Earth observation.

\section{Data Collection and Preprocessing}

This study integrates pan-Greenland, multi-modal Earth observation, and reanalysis datasets to monitor and classify supraglacial lake evolution across the Greenland Ice Sheet. Our analysis spans two climatologically distinct melt seasons—2018 and 2019, which together provide a valuable contrast: 2018 was characterized by relatively cooler and less intense surface melting, while 2019 experienced anomalously warm conditions and widespread surface melt. This temporal range allows us to evaluate model robustness across contrasting climate regimes. Spatially, the dataset encompasses all six major drainage basins of Greenland—Central West (CW), Northeast (NE), North (NO), Northwest (NW), Southeast (SE), and Southwest (SW)—ensuring comprehensive coverage of the diverse glaciological and topographic environments represented in prior ice sheet studies \cite{howat2014greenland}.

\begin{figure*}[t]
    \centering
    \includegraphics[width=0.75\linewidth]{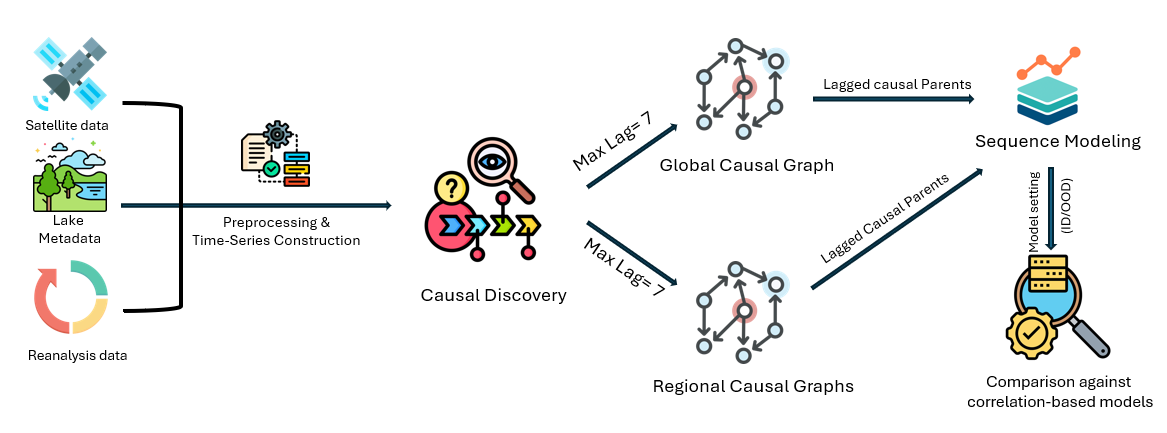}
    \caption{
    Overview of the proposed causally-informed classification framework for supraglacial lake evolution. Multimodal remote sensing and reanalysis data are processed into daily per-lake time series. Using J-PCMCI+, we identify time-lagged causal relationships that reveal robust predictors of lake dynamics, both globally and by region. These causal features are then used in MiniROCKET-based sequence modeling to classify lake outcomes. The framework is evaluated under both in-distribution and out-of-distribution settings to assess generalization under spatial distribution shift.
    }
    \label{fig:methodology_pipeline}
\end{figure*}

\textbf{(1) Sentinel-1 Synthetic Aperture Radar (SAR):}  
We use Sentinel-1 Level-1 Ground Range Detected (GRD) SAR imagery, acquired in Interferometric Wide (IW) swath mode and preprocessed via Google Earth Engine (GEE), which includes thermal noise removal, radiometric calibration, terrain correction, and log-scaling to decibels. The product provides C-band backscatter at $\sim$10-meter resolution with a nominal revisit frequency of 6 days during 2018 and 2019 due to the combined Sentinel-1A/B constellation.

We extract horizontally-transmitted, vertically-received (HV) polarization values, which are sensitive to surface roughness and dielectric contrast, making them ideal for identifying liquid water on ice \cite{nagler2015sentinel}. For each lake, we compute the average HV backscatter within the lake boundary (HV\textsubscript{lake}) and in a 750-meter buffer outside the lake (HV\textsubscript{out}). We then calculate the backscatter anomaly using Equation~(\ref{eq:hv_anom}).
This contrast normalizes for orbit differences and improves the detection of water presence within lakes. The time series of HV\textsubscript{anom} is linearly interpolated and smoothed using a 12-day rolling median to ensure temporal continuity.

\textbf{(2) Sentinel-2 and Landsat-8 Optical Imagery:}  
We use top-of-atmosphere reflectance imagery from Sentinel-2 Level-1C and Landsat-8 Collection 1 Tier 1, accessed via GEE. To compute water fraction and solar zenith angle metrics, we use the following bands:
\begin{itemize}
    \item \textbf{Sentinel-2:} Band 2 (Blue, 20 m), Band 3 (Green, 20 m), Band 4 (Red, 20 m), Band 10 (Cirrus, 60 m), Band 11 (SWIR 1, 20 m)
    \item \textbf{Landsat-8:} Band 2 (Blue, 30 m), Band 3 (Green, 30 m), Band 4 (Red, 30 m), Band 6 (SWIR 1, 30 m)
\end{itemize}

We apply cloud and shadow masking using thresholds on SWIR and Cirrus bands (e.g., SWIR $>$ 0.1 or Cirrus $>$ 0.1 for S2), and identify water pixels using the Normalized Difference Water Index (NDWI), retaining pixels with NDWI $>$ 0.18 for Sentinel-2 and NDWI $>$ 0.19 for Landsat-8 \cite{moussavi2020antarctic}. From each valid image, we compute:
\begin{itemize}
    \item S2\textsubscript{water} and LS\textsubscript{water}: Percentage of cloud-free water pixels within the lake polygon.
    \item S2\textsubscript{zenith} and LS\textsubscript{zenith}: Mean solar zenith angle across lake bounds.
\end{itemize}

Outlier observations caused by shadow or poor illumination are removed using temporal filtering rules based on surrounding values and solar zenith angles. We denote the optical water-fraction generically as $p_{\text{water}}$; when the sensor matters, we write S2\textsubscript{water} or LS\textsubscript{water}. Finally, we interpolate the water fraction time series to generate daily values and apply a 12-day median filter.

\textbf{(3) CARRA Climate Reanalysis:}  
We incorporate climate data from the Copernicus Arctic Regional ReAnalysis (CARRA-West) \cite{schyberg2020carra}, which provides 3-hourly meteorological fields over the Greenland Ice Sheet at a 2.5~km spatial resolution. For each lake centroid, we extract daily-averaged values from the nearest CARRA grid cell for the following:

\begin{itemize}
    \item \textbf{t2m} – Near-surface (2~m) air temperature (K)
    \item \textbf{r2} – Relative humidity at 2~m (\%)
    \item \textbf{sp} – Surface pressure (Pa)
    \item \textbf{sst} – Sea surface temperature (K)
\end{itemize}

We apply linear interpolation to fill short gaps and smooth the series using a 12-day rolling median. These variables provide atmospheric context relevant for lake evolution.

\textbf{(4) Supraglacial Lake Metadata:}  
We utilize lake outlines and attributes from the pan-Greenland inventory introduced by \cite{dunmire2024greenland}, which includes 3,846 and 6,146 lakes detected in 2018 and 2019, respectively. Polygons are digitized at 30-meter resolution from optical imagery and contain metadata fields: \textit{lake\_id}, \textit{region}, \textit{label} (refreeze, buried, rapid drainage, slow drainage), \textit{area} (m\textsuperscript{2}), \textit{elevation} (m), and \textit{year}. We use a manually labeled subset of 1,000 lakes, with 250 per class, to ensure balanced representation across evolution types. These annotations serve as ground-truth labels for supervised classification tasks.

All data sources are temporally aligned on a per-lake basis and resampled to generate complete 365-day time series per lake per year, enabling consistent and temporally rich input for causal discovery and downstream modeling.

\section{Discovering Temporal Causal Graphs}

To build models of supraglacial lake evolution that are both robust and scientifically grounded, it is not enough to rely on correlation-based features. What matters is identifying which variables are truly causal and at what temporal delays they influence lake dynamics. Our contribution here is to move beyond simple feature selection by explicitly discovering and incorporating \textit{causal variables together with their lag structures}. This ensures that downstream models capture not only which features matter, but also when their effects occur.

We use the J-PCMCI+ algorithm for causal discovery over multivariate time series constructed from satellite imagery, climate reanalysis, and lake metadata. J-PCMCI+ is a constraint-based method for multi-domain causal inference that uncovers stable, time-lagged relationships while controlling for spatial and temporal heterogeneity via structured dummy variables~\cite{runge2020discovering}. Each lake is treated as an observational unit with a 365-day multivariate time series. 

Causal discovery is performed in two settings: 

\begin{itemize}
    \item \textbf{Global causal graph:} Estimated from 60 lakes (10 per region) across all six drainage basins, capturing causal patterns that transfer across Greenland.
    \item \textbf{Region-specific causal graphs:} Estimated independently for each basin using 20 lakes, capturing localized drivers that differ across glaciological settings.
\end{itemize}

We adopt a maximum lag of 7 days with significance level $\alpha = 0.01$, using the RegressionCI estimator suited for continuous, autocorrelated variables. This short lag window reflects the synoptic to weekly memory of surface processes controlling lakes. Longer windows would add noise and reduce physical interpretability. 

The input time series includes radar backscatter anomaly (HV\textsubscript{anom}), optical water fraction and solar zenith from Sentinel-2 and Landsat-8 (S2\textsubscript{water}, LS\textsubscript{water}, S2\textsubscript{zenith}, LS\textsubscript{zenith}), and meteorological variables from CARRA (t2m, r2, sp, sst). Three structured dummies encode context:
\begin{itemize}
    \item \textbf{s-dummy}: lake identity (intra-region variation),
    \item \textbf{r-dummy}: basin (inter-region variation),
    \item \textbf{t-dummy}: day-of-year (seasonal phase).
\end{itemize}
These reduce spurious links tied to spatial or seasonal labels.

The discovered graphs highlight temporal drivers of lake evolution. HV\textsubscript{anom} shows strong autoregression, reflecting persistence in radar backscatter. Optical water fraction (S2\textsubscript{water}, LS\textsubscript{water}) is consistently identified at lag 0, confirming the immediate role of surface water. Region-specific graphs further reveal localized predictors such as r2 (relative humidity) in CW and sst (sea surface temperature) in NW. Due to space limits, full graphs are provided in the GitHub repository.\footnote{\url{https://github.com/ehfahad/RIC-TSC}}

\begin{table}[!h]
\centering
\caption{Discovered causal parents of HV\textsubscript{anom} and their lags using J-PCMCI+. Variables span satellite observations, meteorological drivers, and structured dummies.}
\label{tab:causal_parents}
\scriptsize
\begin{tabular}{c p{6.6cm}}
\hline
\textbf{Region} & \textbf{Causal Parents (time lags)} \\
\hline
ALL & HV\textsubscript{anom}(-1,-2,-3,-4,-5,-7); S2\textsubscript{water}(0,-1,-2); r-dummy(0) \\
CW & HV\textsubscript{anom}(-1,-2,-3,-4); S2\textsubscript{water}(0); r2(0) \\
NE & HV\textsubscript{anom}(-1,-2,-3,-4,-5); t-dummy(0) \\
NO & HV\textsubscript{anom}(-1,-2,-3,-4,-7); LS\textsubscript{water}(0); S2\textsubscript{water}(0) \\
NW & HV\textsubscript{anom}(-1,-2,-3,-4,-7); LS\textsubscript{water}(0); S2\textsubscript{water}(0); sst(0); s-dummy(0) \\
SE & HV\textsubscript{anom}(-1,-2,-3,-4); LS\textsubscript{water}(0) \\
SW & HV\textsubscript{anom}(-1,-2,-3,-4); S2\textsubscript{water}(0); LS\textsubscript{water}(0); t-dummy(0) \\
\hline
\end{tabular}
\end{table}

Table~\ref{tab:causal_parents} lists the causal parents of HV\textsubscript{anom}. The global graph captures Greenland-wide trends, while regional graphs expose differences in local forcing mechanisms. The key contribution is not merely variable selection but the explicit inclusion of lagged causal predictors. These are passed into the MiniROCKET model, allowing it to learn from temporally grounded signals rather than arbitrary correlations. This reduces dimensionality, preserves causal structure, and improves robustness under distribution shifts. The next section evaluates these causal inputs in classification experiments under both in-distribution and out-of-distribution settings.

\section{Experimental Setup}

The objective of this study is to evaluate whether causally-informed time series modeling improves the robustness and generalization of supraglacial lake evolution prediction under spatial and climatic variability. Each lake is represented as a 365-day multivariate time series comprising satellite-derived observations, climate reanalysis variables, and contextual metadata. The task is framed as a four-class classification problem with labels \textit{refreeze}, \textit{buried}, \textit{slow drainage}, and \textit{rapid drainage}. These categories were defined from expert-validated interpretations of radar and optical time series, reflecting distinct meltwater dynamics that influence surface hydrology, firn structure, and potential sea-level contribution.

\subsection{Modeling Framework}

We adopt a lightweight but expressive architecture designed for generalizable scientific time series learning. Each lake’s daily record is first encoded using \textbf{MiniROCKET}~\cite{dempster2021minirocket}, a deterministic convolutional transform that produces high-dimensional, discriminative feature vectors. These are classified with a \textbf{RidgeClassifier}, where the regularization parameter is selected via internal cross-validation.

To isolate the role of causal discovery, we compare two pipeline variants:

\begin{itemize}
    \item \textbf{Causal Model:} Uses only the lagged causal parents of HV\textsubscript{anom} identified by J-PCMCI+. Parents are drawn from global or region-specific graphs and include statistically validated predictors with a maximum lag of seven days, providing a compact and causally-grounded feature set.
    
    \item \textbf{Correlation-Based Baseline:} Uses all nine observed variables (e.g., HV\textsubscript{anom}, S2\textsubscript{water}, t2m) without causal filtering or lag constraints, reflecting the common practice in Earth observation of feature selection by availability or correlation.
\end{itemize}

This design enables a direct comparison between causally-informed and correlation-based models in terms of generalization, interpretability, and parsimony.

\subsection{Evaluation Protocol}

We evaluate model robustness and cross-domain generalization under three scenarios:

\paragraph{Global Evaluation (Pooled Data).}  
An 80/20 stratified split is applied to the full dataset pooled across all six regions. This benchmark measures aggregate performance under mixed spatial and seasonal conditions.

\paragraph{Region-Specific Evaluation (In-Distribution).}  
For each of Greenland’s six drainage basins, we conduct an 80/20 stratified split within the region. This assesses whether region-specific causal graphs improve prediction when training and testing occur in the same geophysical domain.

\paragraph{Region-Held-Out Evaluation (Out-of-Distribution).}  
We implement a leave-one-region-out strategy, training on a single basin and testing on the remaining five. This simulates deployment to unseen geophysical and climatological regimes.

Performance metrics are chosen to match the objectives of each setting. In the global experiment, we report multi-class accuracy together with macro-averaged precision, recall, and F1-score to ensure a balanced evaluation across classes. For the in-distribution and out-of-distribution settings, we report accuracy alone. Since the dataset is perfectly balanced (250 lakes per class), accuracy provides a sufficient and interpretable measure of model performance, while additional metrics would not provide new insight. All data and code used for these experiments are available in the accompanying GitHub repository. Across these evaluations, our framework demonstrates the value of integrating causally discovered predictors and their lagged effects into sequence modeling, going beyond correlation-based baselines to improve robustness, interpretability, and generalization under distribution shifts.

\section{Results and Discussion}

We evaluate the proposed framework across three regimes: (i) global classification on pooled data, (ii) region-specific classification under in-distribution (ID) settings, and (iii) region-held-out generalization under out-of-distribution (OOD) shift. These settings test whether causally-informed modeling with time-lagged drivers improves prediction of supraglacial lake evolution compared to correlation-based baselines.

The causal model leverages only statistically validated lagged parents of HV\textsubscript{anom}, whereas the baseline uses all observed variables without filtering. This isolates the effect of causal discovery and highlights the role of domain-consistent predictors in improving generalization and interpretability.

\subsection{Global Classification}

\begin{figure}[!h]
    \centering
    \includegraphics[width=0.85\linewidth]{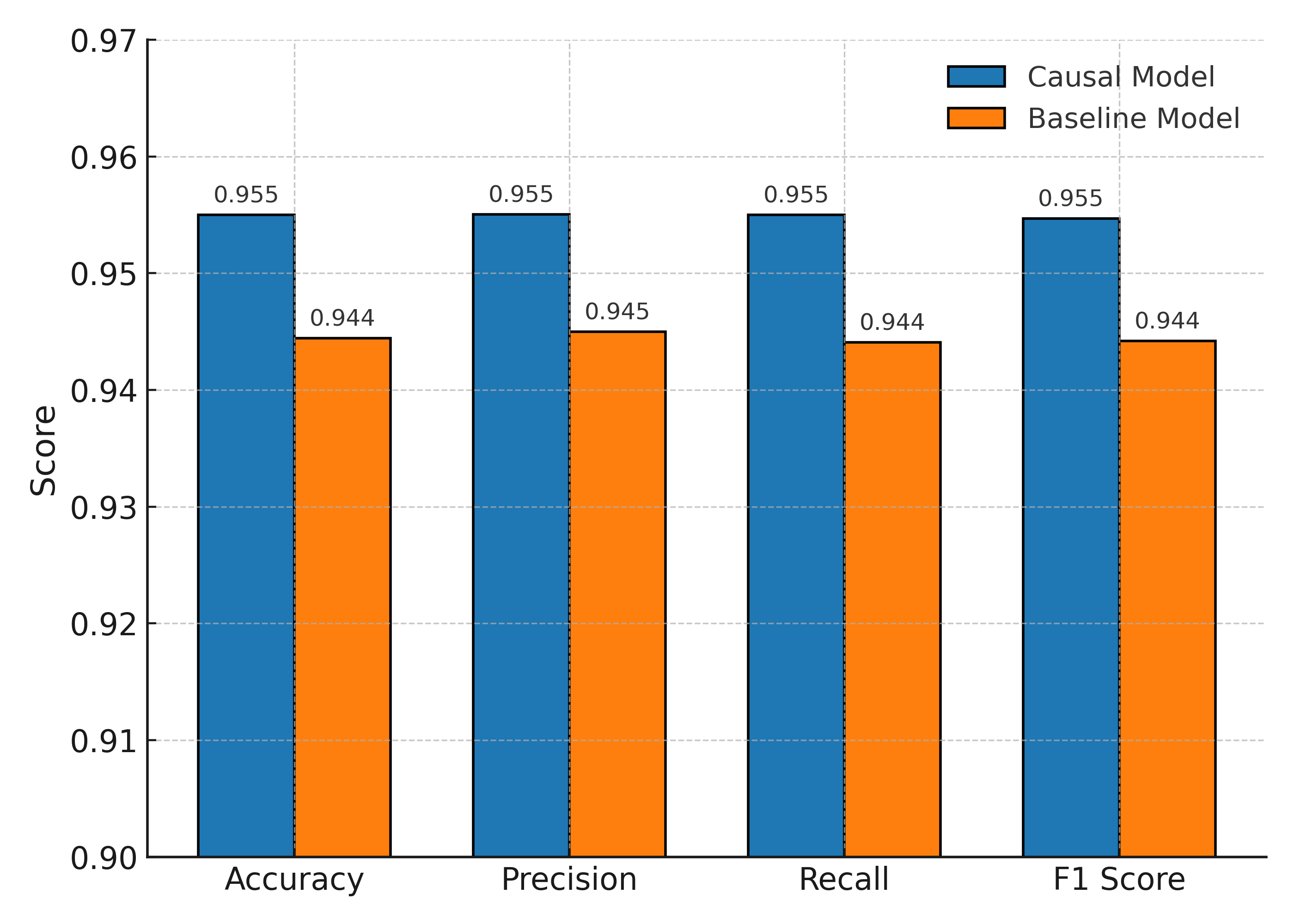}
    \caption{Global classification performance using accuracy, macro-averaged precision, recall, and F1-score.}
    \label{fig:global-bar}
\end{figure}

On the pooled dataset, an 80/20 stratified split ensures balanced representation across lake types and regions. Figure~\ref{fig:global-bar} reports performance using accuracy, precision, recall, and F1-score.

The causal model achieves higher accuracy (+1.06\%) and consistent gains across all metrics. Notably, these improvements stem from a reduced feature set composed of lagged causal predictors, showing that filtering enhances efficiency and emphasizes physically meaningful signals. Even without an explicit distribution shift, this provides more interpretable and compact models.

\subsection{Region-Specific Classification}

We next evaluate models trained and tested independently within each drainage basin (Table~\ref{tab:region-id}). This setup measures how region-specific causal graphs perform under localized conditions.

\begin{table}[!h]
\caption{Region-wise in-distribution (ID) classification accuracy. Bolded values indicate positive accuracy gain of the causal model.}
\label{tab:region-id}
\centering
\scriptsize
\begin{tabular}{lccc}
\toprule
\textbf{Region} & \textbf{Causal Model} & \textbf{Baseline Model} & \textbf{Accuracy Gain} \\
\midrule
CW & 88.10\% & 87.80\% & \textbf{+0.29\%} \\
NE & 78.79\% & 84.38\% & --5.59\% \\
NO & 89.29\% & 85.71\% & \textbf{+3.57\%} \\
NW & 90.32\% & 90.32\% & 0.00\% \\
SE & 78.95\% & 78.95\% & 0.00\% \\
SW & 91.84\% & 89.80\% & \textbf{+2.04\%} \\
\bottomrule
\end{tabular}
\end{table}

Causal features improve or match baseline performance in four of six regions, with the largest gains in North (NO) and Southwest (SW). The Northeast (NE) shows a decline, likely reflecting noise and data sparsity, highlighting the importance of uncertainty handling in discovery. Overall, region-specific graphs capture local drivers and offer accuracy gains where dynamics diverge from global trends. Importantly, these results demonstrate that causal features are not only competitive with baselines but can reveal regional heterogeneity that global models may overlook.

\subsection{Region-Held-Out Classification}

The most demanding evaluation tests transferability under spatial distribution shift via leave-one-region-out (LOO). Table~\ref{tab:region-ood} summarizes results.

\begin{table}[!h]
\caption{Region-wise out-of-distribution (OOD) classification accuracy. Bolded values indicate positive accuracy gain of the causal model.}
\label{tab:region-ood}
\centering
\scriptsize   
\begin{tabular}{lccc}
\toprule
\textbf{Train Region} & \textbf{Causal Model} & \textbf{Baseline Model} & \textbf{Accuracy Gain} \\
\midrule
CW & 89.14\% & 87.15\% & \textbf{+1.99\%} \\
NE & 89.99\% & 83.39\% & \textbf{+6.59\%} \\
NO & 86.15\% & 73.56\% & \textbf{+12.59\%} \\
NW & 88.31\% & 81.84\% & \textbf{+6.47\%} \\
SE & 83.04\% & 77.20\% & \textbf{+5.84\%} \\
SW & 90.07\% & 86.73\% & \textbf{+3.34\%} \\
\bottomrule
\end{tabular}
\end{table}

The causal model consistently outperforms the baseline across all held-out regions, with gains up to +12.59\%. These improvements show that lagged causal drivers provide temporally stable and physically grounded signals that generalize across distinct glaciological settings. By filtering spurious correlations, the model preserves predictive fidelity in unseen domains.

\subsection{Discussion}

Together, these results show that causally-informed time series modeling improves both accuracy and robustness for supraglacial lake evolution classification. The gains are modest in pooled and ID settings but substantial under OOD shift, where generalization is most challenging. This supports our design choice to incorporate causal lags alongside efficient sequence modeling.

The framework further balances interpretability and efficiency: discovery reduces dimensionality while retaining predictors aligned with known processes, and the lightweight MiniROCKET–Ridge pipeline avoids overfitting. While future comparisons with deep sequence models (e.g., LSTMs, Transformers) could broaden context, our results establish causally-informed lag selection as a practical and scientifically grounded approach for spatiotemporal Earth observation tasks.

\section{Conclusion and Future Work}

This work introduces a regionally-informed framework, RIC-TSC, for modeling supraglacial lake (SGL) evolution that integrates Joint PCMCI+ (J-PCMCI+) with MiniROCKET in a unified pipeline. Instead of treating causal discovery as a preprocessing step, we directly embed statistically validated lagged predictors into time series classification. The resulting models are lightweight, interpretable, and achieve up to 12.59\% higher accuracy than correlation-based baselines under cross-region generalization. Crucially, causal drivers improve robustness to distribution shift while anchoring predictions in known glaciological processes.

Our central contribution is to show that causal discovery enables more than dimensionality reduction: it uncovers invariant and region-specific temporal mechanisms governing lake evolution. By distinguishing immediate from delayed influences of radar and optical signals and revealing how climate variables such as relative humidity or sea surface temperature affect basin-level dynamics, the framework links causal learning with physical understanding. This dual focus on predictive performance and mechanistic insight advances Earth observation toward models that generalize across heterogeneous environments while remaining scientifically interpretable.

Nonetheless, challenges remain. J-PCMCI+ is sensitive to sample size and autocorrelation, and expert-derived lake labels may not capture all hydrological subtleties. Future work will stratify causal discovery by lake outcome to isolate class-specific mechanisms and extend the pipeline with spatiotemporal graph neural networks to capture inter-lake dependencies. Embedding causal discovery within scalable classification thus provides a promising path toward generalizable, interpretable, and scientifically grounded modeling of dynamic Earth surface processes under climate change.

\section*{Acknowledgement}
This work is supported by iHARP: NSF HDR Institute for Harnessing Data and Model Revolution in the Polar Regions (Award\# 2118285). The views expressed in this work do not necessarily reflect the policies of the NSF, and endorsement by the Federal Government should not be inferred.

\bibliographystyle{IEEEtran}
\bibliography{References}

\end{document}